%% file: main.tex
\documentclass[journal]{IEEEtran}

\usepackage{color}
\usepackage{bm}
\usepackage{mathtools}

\ifCLASSINFOpdf

\usepackage{graphicx}
\usepackage{siunitx}
\usepackage{hyperref}
\usepackage{amsmath,nccmath}
\usepackage{mathtools}
\usepackage{textcomp,gensymb}

\begin{document}

\newcommand{\rob}[1]
{\textbf{\textcolor[rgb]{0.00,0.80,0.00}{[Rob::~#1]}}}
\newcommand{\jun}[1]
{\textbf{\textcolor[rgb]{0.00,0.50,0.50}{[Jun::~#1]}}}
\newcommand{\charlie}[1]
{\textbf{\textcolor[rgb]{0.80,0.00,0.00}{[Charlie::~#1]}}}

\newcommand{\revise}[2]{\textcolor{blue}{#2}}

\title{Sensing and Reconstruction of 3D Deformation on Pneumatic Soft Robots}

\author{Rob~B.N.~Scharff,~\IEEEmembership{Student Member,~IEEE,}
        Guoxin~Fang,~\IEEEmembership{Student Member,~IEEE,} Yingjun~Tian, Jun~Wu, Jo~M.P.~Geraedts, and Charlie~C.L.~Wang$^\dag$,~\IEEEmembership{Senior Member,~IEEE}
\thanks{R.B.N.~Scharff, G.~Fang, J.~Wu, and J.M.P.~Geraedts are with the Department of Sustainable Design Engineering, Delft University of Technology (TU Delft), The Netherlands.} 
\thanks{Y.~Tian is with the Department of Mechanical and Automation Engineering, The Chinese University of Hong Kong, China} 
\thanks{C.C.L.~Wang is with the Department of Mechanical, Aerospace and Civil Engineering, The University of Manchester, UK} 
\thanks{$^\dag$Corresponding Author: {\tt \small changling.wang@manchester.ac.uk} }
\thanks{Manuscript prepared on December 18th, 2020.}
}

\markboth{}
{Scharff \MakeLowercase{\textit{et al.}}: Sensing and Reconstruction of 3D Deformation on Pneumatic Soft Robots }

\maketitle

\begin{abstract}
Real-time proprioception is a challenging problem for soft robots, which have almost infinite degrees-of-freedom in body deformation. When multiple actuators are used, it becomes more difficult as deformation can also occur on actuators caused by interaction between each other. To tackle this problem, we present a method in this paper to sense and reconstruct 3D deformation on pneumatic soft robots by first integrating multiple low-cost sensors inside the chambers of pneumatic actuators and then using machine learning to convert the captured signals into shape parameters of soft robots. An exterior motion capture system is employed to generate the datasets for both training and testing. With the help of good shape parameterization, the 3D shape of a soft robot can be accurately reconstructed from signals obtained from multiple sensors. We demonstrate the effectiveness of this approach on two designs of soft robots -- a robotic joint and a deformable membrane. After parameterizing the deformation of these soft robots into compact shape parameters, we can effectively train the neural networks to reconstruct the 3D deformation from the sensor signals. The sensing and shape prediction pipeline can run at 50Hz in real-time on a consumer-level device.
\end{abstract}

\begin{IEEEkeywords}
Proprioception, 3D Deformation, Pneumatic Actuators, Soft Robotics
\end{IEEEkeywords}

\IEEEpeerreviewmaketitle

\input{secIntroduction.tex}

\input{secCapturing3DShape.tex}

\input{secSoftRobotRealization.tex}

\input{secResults.tex}

\input{Discussion.tex}

\section*{Acknowledgment}
R.B.N. Scharff's research is supported by the Faculty of Industrial Design Engineering at TU Delft. This project is also financially supported by the Department of Mechanical and Automation Engineering, the Chinese University of Hong Kong and the CUHK Technology and Business Development Fund (ref. no.: TBF19ENG005). 

\ifCLASSOPTIONcaptionsoff
  \newpage
\fi

\bibliographystyle{IEEEtran}
\bibliography{References.bib}

\vfill

\end{document}

%% file: secIntroduction.tex
\section{Introduction}
\label{sec:Introduction}
Proprioception for soft robots is a challenging problem because of the almost infinite \textit{Degrees-Of-Freedom} (DOFs) of the deformable bodies, and because there is no off-the-shelf sensor available. However, accurate proprioception is crucial for closing the loop of control. Existing solutions usually conduct a simplified model according to a specific design of soft robots -- e.g., sensing a single bending angle~\cite{Elgeneidy2018embeddedflexsensors} or curvature~\cite{Zhao2016}. A general and easy-to-fabricate solution for sensing 3D deformation is needed. In this paper, we propose a method using low-cost sensors to realize accurate proprioception and real-time 3D shape reconstruction. Our approach is based on a data-driven strategy that can be generally applied to different designs based on their own shape parameterization. 

\subsection{Related Work}
The literature is reviewed from three angles, namely sensors, deformation acquisition and machine learning.

\subsubsection{Sensors for proprioception} 
A large variety of sensors have been developed for proprioception in soft robotics. For soft bending actuators, proprioceptive sensing is commonly achieved by embedding paths of conductive materials that change their resistivity upon deformation, such as liquid metal \cite{Morrow2016}, a 3D-printed carbon black/PLA compound \cite{Elgeneidy2018printstrainsensors,Yang2018embeddedsensors}, commercial flex sensors based on conductive ink \cite{Elgeneidy2018embeddedflexsensors}, EMIM-ES ionogel \cite{Truby2018}, PDMS impregnated with carbon nanotubes \cite{Thuruthel2019}, or laser-cut patterns from off-the-shelf conductive silicone \cite{Truby2020}. Proprioception can also be achieved by magnetic sensing \cite{Ozel2016} and inductance-based sensing \cite{Felt2016}. The inductance-based sensing method can also be applied to a continuum joint~\cite{Felt2019}. However, most of the sensors mentioned above cannot accommodate very large strains or cannot capture multiple DOFs, which makes them unsuitable for other types of actuators such as elongational actuators or 3D deforming surfaces. Moreover, integrating these sensors into an actuator is usually cumbersome during fabrication.

The use of optical sensing for proprioception in soft robots has shown to have great potential. Examples of optical sensing for soft robots include the use of stretchable optical waveguides for use in bending actuators~\cite{Zhao2016,Bai2020}, macrobend optical sensing for elongational actuators~\cite{Sareh2015}, optical distance sensors on a helical flexible printed circuit board for a soft robotic joint~\cite{Teichert2019}, fiber optics in a three-dimensionally deforming surface~\cite{Wang2019}, the use of fluidic channels in combination with an external camera~\cite{Soter2020}, and embedded cameras for tactile sensing~\cite{Ward2018, Wiertlewski2020, Yuan2017gelsight} and inflatable bellows~\cite{Werner2020}. However, some of these approaches can only sense relative simple deformation (e.g., it is difficult to embed optical waveguides and fluidic channels inside 3D deforming surfaces or elongational actuators). On the other aspect, the image-texture based methods can only be used in large actuators according to the size of cameras. Color sensors have been used in \cite{Scharff2018,Scharff2019} to increase the accuracy of proprioception; however, the fabrication of actuators with color patterns needs to use high-end 3D printers, which is demanding. Differently, in this paper we propose a general sensing strategy and demonstrate its use in both bending actuators of various sizes~\cite{Scharff2018,Scharff2019} and elongational actuators~\cite{Scharff7}.

\subsubsection{Deformation acquisition}
An important challenge in sensing soft robot deformation is how to obtain the accurate ground truth information in deformation. Simplified information is sensed in the prior researches, including the bending angle~\cite{Elgeneidy2018embeddedflexsensors}, the curvature~\cite{Zhao2016} and the position of tip point~\cite{Giorelli2015,vanMeerbeek2018}. However, important information on the shape of a soft robot is lost. A straightforward solution is to increase the number of sensed points on the actuator. However, the number of points is limited when physical manipulators~\cite{vanMeerbeek2018} or sensors (e.g. inertial measurement units \cite{Felt2019} or electromagnetic sensors~\cite{Sareh2015,Xu2013}) are used to determine the position of each point. 
For these reasons, systems that capture markers on a soft robot with one or more cameras are a popular choice for capturing ground truth information of soft objects (ref.~\cite{Felt2019,Wang2019,Scharff2019,Bacher2016,Lun2019}). The captured marker coordinates on the robot can be used to reconstruct the complete shape of a soft robot \cite{Glauser2019}. Therefore, it is used in our work to capture the ground truth information for soft actuators with different types of deformation.

\subsubsection{Machine learning}
Due to the highly nonlinear deformation presented on soft bodies, it is hard to build an accurate analytical sensing model for soft robots. Simplified analytical models can only be applied to specific type of design -- i.e., do not have generality. Machine learning methods, in particular artificial neural networks, have shown to be a powerful tool to learn these nonlinearities while being applicable to a wide range of designs. \textit{Feedforward Neural Networks} (FNNs) have been used to learn the kinematics of soft robots~\cite{Giorelli2015, Runge2017} and to characterize various types of soft sensors~\cite{Scharff2019,vanMeerbeek2018,Glauser2019,Lun2019,Sferrazza2018}. 
When sequential data is collected, a \textit{Recurrent Neural Network} (RNN) or \textit{Long Short Term Memory} network (LSTM) can be used to include time-variant effects such as hysteresis in the sensing model~\cite{Thuruthel2019,Truby2020,Soter2020,Han2018}. As a powerful tool when working with camera data as sensor input, \textit{Convolutional Neural Network} (CNN) has been employed in combination with an LSTM to calibrate a soft tactile sensor for detecting the hardness of objects \cite{Yuan2017hardnessestimation}. We employ neural networks in our learning process to establish the mapping between the signals from sensors and the shape parameters that are extracted from the captured positions of markers.

\begin{figure}[t]
\centering
\includegraphics[width=\linewidth]{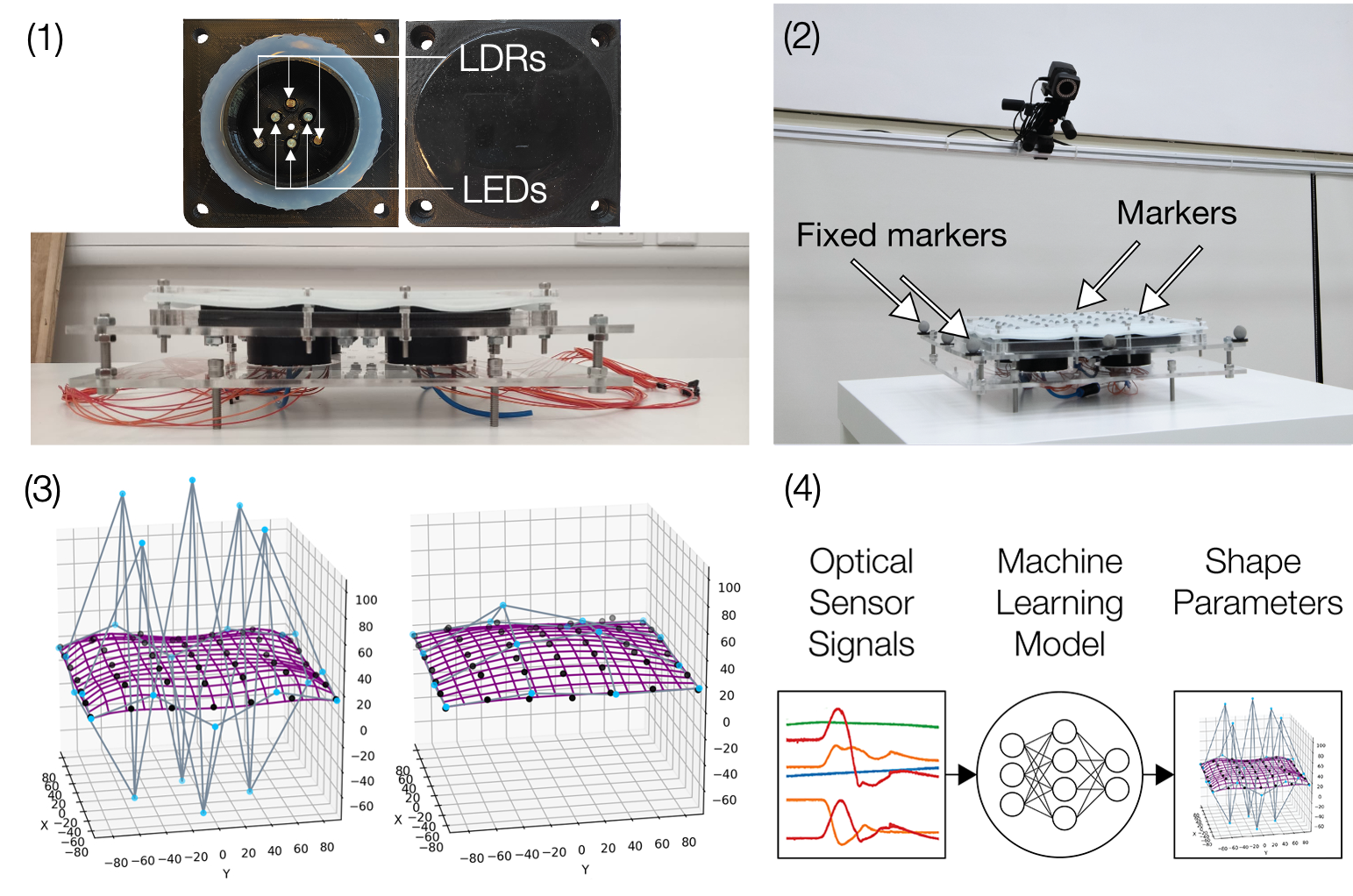}
\caption{Overview of our approach for enabling the sensing capability of 3D shape on soft robots by four steps: (1) sensor integration, (2) data collection, (3) shape parameterization and (4) learning the mapping.}\label{fig:appproach}
\end{figure}

\subsection{Our Approach}
Our approach endows soft pneumatic actuators with the sensing capability for real-time 3D shape reconstruction by four steps (see Fig.~\ref{fig:appproach}):
\begin{enumerate}
\item Embedding optical sensors and lamps into the air chambers of soft robots to translate deformations of the air chambers to the variation of light intensity;

\item Capturing the ground truth deformation of the soft robot using markers placed on the robot;

\item Extracting shape parameters, that can represent deformation more compactly, from the positions of markers;

\item Learning the mapping between the signals captured on sensors and the corresponding deformation represented by shape parameters. 
\end{enumerate}

In short, the technical contributions of our work are:
\begin{itemize}
    \item Accurate proprioception for pneumatic soft robots consisting of multiple interacted actuators undergoing 3D deformation;
    
    \item Capability to reconstruct a 3D deformed shape by learning the mapping between the sensed signals and the shape parameters.
\end{itemize}
Both the sensing and the reconstruction steps can run at 50Hz on a consumer-level device. Two robot designs -- a robotic joint and a deformable membrane are selected to demonstrate these new capabilities, which are considered as essential advancement after developing the capability of proprioception on a soft pneumatic bending actuator~\cite{Scharff2018,Scharff2019} and on a linear bellow~\cite{Scharff7}.

%% file: secCapturing3DShape.tex
\section{Sensing Deformation on Soft Actuators}
\label{sec:3Dshape}
This section serves to explain the importance of using multiple sensors to capture the deformation of pneumatic actuators on soft robots. Deformation of an actuator with one DOF can often be captured by a single sensor. However, the signals captured by a single sensor can hardly distinguish the configurations of deformation in multi-DOFs. Attempts have been made to increase the number of signals that can be captured by using a camera instead of single (optical) sensors (ref.~\cite{Soter2020,Ward2018, Wiertlewski2020, Yuan2017gelsight,Werner2020}). For example if images are taken at the resolution of $1280 \times 720$, this method can capture up to  $ 1280 \times 720 \times 3 = 2,764,800$ different signals. However, in practice, the number of sensors required for capturing 3D deformation is much smaller than this -- i.e., redundancy exists in sensing and computation. Besides the waste in computing time, another downside of camera-based sensing is its difficulty to integrate into a narrow space which is quite common in many soft actuators. In our approach, we place a few \textit{Light-Emitting Diodes} (LEDs) and \textit{Light-Dependent Resistors} (LDRs) inside each air chamber to capture the deformation inside a chamber. The signals captured in all chambers are later fused to reconstruct 3D shape of the soft robot driven by these chambers. 

\begin{figure}[t]
\centering
\includegraphics[width=\columnwidth]{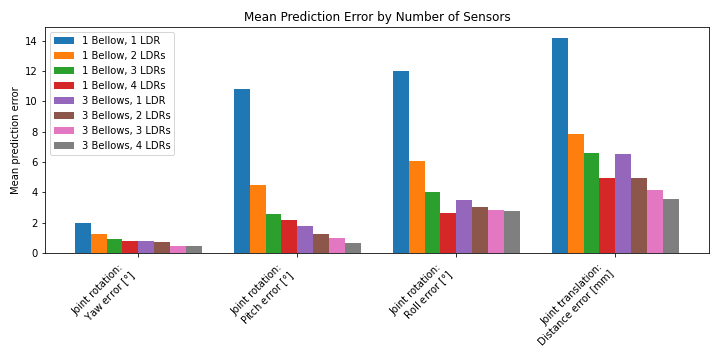}
\includegraphics[width=\columnwidth]{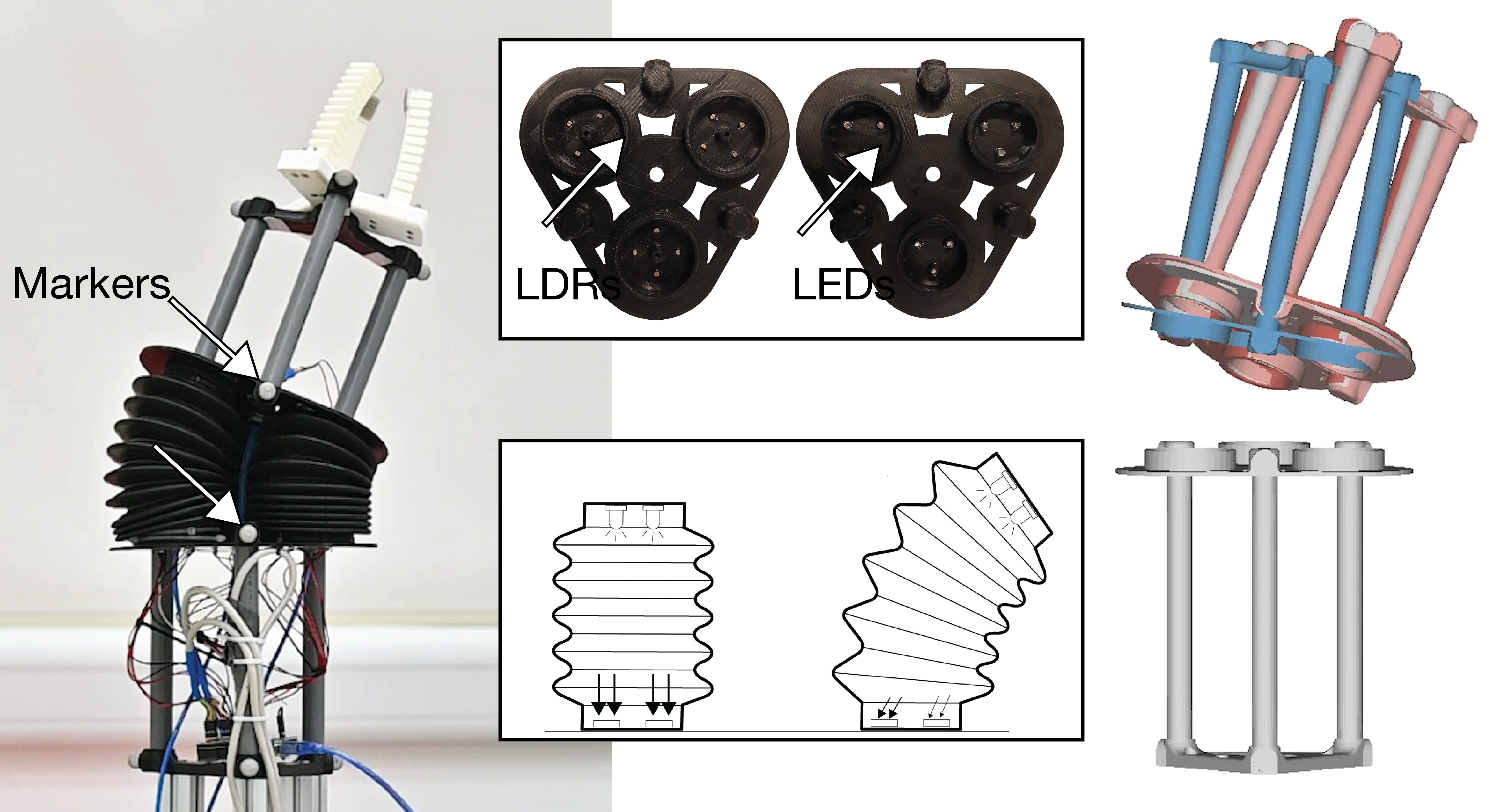}
\caption{
An experiment to demonstrate the importance of sensing the deformation inside one chamber -- (top row) the mean prediction errors generated by using different numbers of LDRs inside either only one bellow or inside all three bellows. (Bottom row) It is found that the predicted transformation is already very accurate when using 4 LDRs in one bellow (light red). For the purpose of comparison, the predictions by using only 1 LDR in one bellow (blue) and 4 LDRs in all three bellows (gray) are also given. The fitted result is indicated in dark red. 
The layout of LDRs and LEDs and the illustration of their working principle are also given in the bottom row. 
}
\label{fig:numberSensors}
\end{figure}

It is important to capture the deformation on each chamber. As shown in the experiment of Fig.\ref{fig:numberSensors}, the 3D transformation can already be well reconstructed even if only one bellow (i.e., the air chamber) is equipped with sensors. However, this chamber should have more sensors -- see the difference between blue (only 1 LDR inside the bellow) and light red (with 4 LDRs inside the bellow). As the ends of the three bellows are connected to the same rigid frames, their deformation are somewhat coupled. Therefore, the accuracy obtained using 4 LDRs in one bellow already approaches that of using 4 LDRs in every bellow (i.e., $4 \times 3 = 12$ LDRs in total). However, this is not the case for many other soft robots with multiple actuators, such as the deformable membrane shown in Figs.\ref{fig:appproach} and \ref{fig:softsurface}.

%% file: secSoftRobotRealization.tex
\section{Soft Robot Realization}
\label{sec:softrobotdesigns}
In this section, we present the realization of our sensing and reconstruction method on two different designs of soft robots. Methods for data acquisition and shape parameterization are also introduced. Lastly, the feasibility of using different machine learning approaches is discussed.

\subsection{Two Designs of Soft Robots}
\subsubsection{Soft continuum joint} 
The design of a soft continuum joint has been indicated in Fig.~\ref{fig:numberSensors}. The soft continuum joint is composed of three inflatable bellows, the top and bottom of which are attached to two rigid frames. Centres of these frames are connected by a cable that constrains the longitudinal expansion of the bellows such that a multi-directional bending motion can be generated upon pressurizing the bellows with different pressures. The top and bottom frame of the joint are fabricated by using \textit{Fused Deposition Modeling} (FDM) with black PLA filaments. The leads of the LEDs and LDRs are fed through small holes in the frames and sealed with epoxy glue. The bellows are off-the-shelf \textit{Freudenberg V6-00400}-bellows \footnote{Freudenberg -- \href{www.fst.com}{www.fst.com}}, which are sealed around the cylindrical parts of the frame using cable ties. In each bellow, 4 LDRs are mounted on the bottom frame to measure the light intensity inside the bellow. This light is generated by 3 LEDs mounted on the top frame of each bellow. Deformations of the bellows result into variation of light intensity that is sensed by the LDRs. This information can be used to determine the deformed shape of a bellow. As different external forces are applied to the joint when the gripper mounted on top of the arm holds different objects, rotation and translation of the joint cannot be determined from the pressure of air inside the bellows. Sensors are needed to determine the rotation and translation of the joint based on the shape variation of the bellows. Due to applying machine learning, an accurate mapping between the sensor signals and soft robot shape can be learned regardless of variations in the sensors or their placement.

\begin{figure}[t]
\centering
\includegraphics[width=\linewidth]{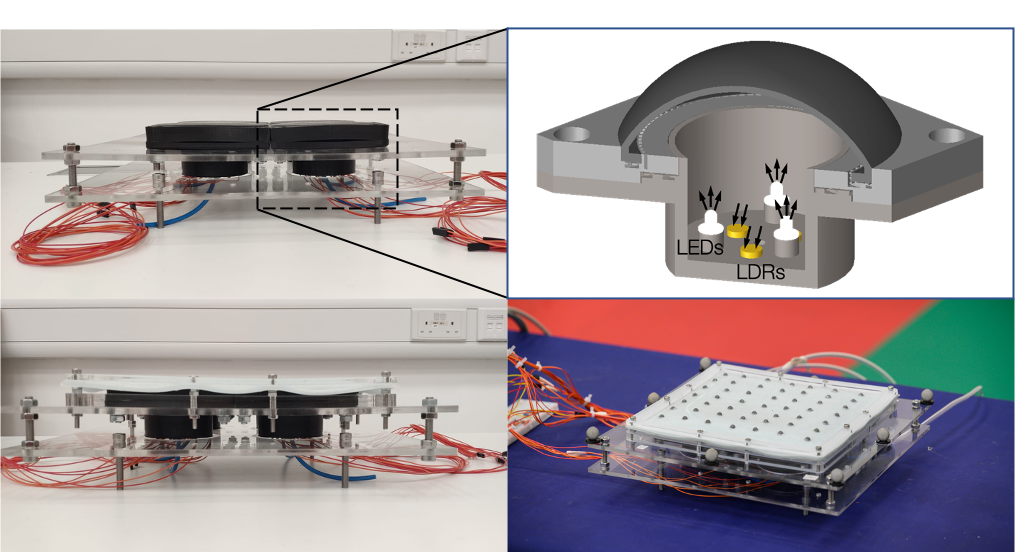}
\caption{The hardware setup of a deformable membrane, which is composed of four inflatable modules. The chamber of each module is mounted with 3 LEDs and 3 LDRs to sense the deformation of each chamber, the signals of which are finally fused to predict the surface shape of the silicone layer. Markers are attached for learning the 3D surface shape by a motion capture system.}\label{fig:softsurface}
\end{figure}

\subsubsection{Soft deformable membrane} 
The design of a deformable membrane is as shown in Fig.~\ref{fig:softsurface}. This hardware setup is composed of four chamber modules that can be inflated separately. The bottom of each module is rigid and mounted with 3 LEDs and 3 LDRs. The chamber is sealed by a lid with a thin inflatable silicone layer. The modules have been fabricated using a combination of FDM and silicone casting. A mechanical interlocking structure, as proposed by Rossing et al.~\cite{Rossing2020}, is used to create an airtight bonding between the 3D printed part and the silicone. The silicone used is \textit{Smooth-On Dragonskin} with a shore hardness of $30A$ colored with black silicone pigment. The filament used is black PLA. The materials are selected for their opacity in order to eliminate the influence of external lighting conditions on the sensor readings. When the chamber of a module is pressurized, the silicone layer inflates. This inflation results into a change in reflection of the light emitted by the LEDs, which is sensed by the LDRs. This information can be used to determine the shape of the inflated silicone. Four modules are mounted on a frame in a $2 \times 2$ layout and covered by a thin layer of silicone to create a smooth deformable membrane. Note that as all the four inflatable modules are interacting with the silicone layer and therefore are coupled, the shape of each module cannot be determined from the air pressure of each chamber. This effect is amplified by the highly non-linear material behavior of the silicone. Therefore, sensors are essential to determine the shape of the membrane.

\subsection{Training Dataset}
This subsection introduces the method of generating the dataset for training. 

\subsubsection{Setup for data acquisition}
A motion capturing system of \textit{Vicon} was used to capture a number of strategically placed markers on the soft robots. For the soft continuum joint, markers were placed at the top and the bottom frame of the joint (seen  Fig.~\ref{fig:numberSensors}). For the soft deformable membrane, a layout of $7 \times 7$ markers was placed on top of the membrane. Additional markers were placed on the rigid frame as reference points for sensing its orientation. These markers have been illustrated in Figs.~\ref{fig:appproach} and \ref{fig:fittedsurfaces}. Upon data collection, the positions of markers were collected at a frequency of $100~\mathrm{Hz}$, whereas data of all the 12 LDRs was collected at a frequency of $1000~\mathrm{Hz}$. The data was synchronized using the \textit{Vicon Lock Sync Box}.

\subsubsection{Sampling strategy}
A good sampling scheme that spans throughout the robot's workspace as well as a wide range of external loads was found crucial to prevent overfitting in data-driven learning. For the soft continuum joint, a range of weights were added on top of the actuator to enable accurate predictions when different external loads were applied to the joint. The weights held by the gripper on top of the arm are varied from $0$ to $500\mathrm{g}$ in steps of $50\mathrm{g}$. A total of $242,131$ samples were collected in $40\mathrm{m}21\mathrm{s}$. Note that each sample means a collection of the markers' positions here. The data collection was divided into three batches. These batches were collected during different times of a day and with an altered orientation and position of the soft robot in the room to guarantee independence of external lighting conditions and the calibration of the motion capturing system. For the deformable membrane, the actuation sequence is varied to ensure that samples can span the entire working space. The data was collected in two batches with varying positions and orientations of the robot as well as varying lighting conditions. A total of $44,403$ samples were collected in $7\mathrm{m}24\mathrm{s}$ for the soft deformable membrane.

\subsubsection{Data Preparation}
Before further processing of the data, captured coordinates of the markers are converted into a more convenient system aligned with the robots. Origin of the soft continuum joint is selected as the center of the bottom triangle of the joint. The $z$-axis is aligned with the triangle's normal pointing upwards, the $y$-axis is pointing from origin towards one of the markers, and the $x$-axis is then defined as orthogonal to these two axes. For the soft deformable membrane, centroid of the fixed markers on the frame is selected as origin. The axes are defined as making $x$- and $y$-axes aligned with the frame's boundary and $z$-axis pointing upwards. 

\subsection{Shape Parameterization}
The most intuitive way to present the shape of a deformed soft robot is to present it by the predicted locations of markers~\cite{Scharff2019}. However, this is redundant in many scenarios. Two shape parameterizations are introduced below for the hardware setups employed in our work, which in general provides a more compact and effective way to reconstruct the shape of deformed soft robots.  

\subsubsection{Soft continuum joint} A parameterization with physical meaning is demonstrated on this hardware setup. The collected coordinates of markers are converted to a rigid transformation represented by a rotation matrix $\mathbf{R}$ together with a translation vector $\mathbf{T}$, which describes the rotation and translation from the bottom triangle of the joint to the top triangle of the joint. For a set of points (i.e., markers) on the bottom triangle denoted as $\{\mathbf{m}_i\}$ and the corresponding set of points on the top triangle as $\{\mathbf{d}_i\}$, the mapping between them can be described as 
\begin{equation}
    \mathbf{d}_i = \mathbf{R} \mathbf{m}_i+\mathbf{T}+\mathbf{v}_i
\end{equation}
with $\mathbf{v}_i$ being a noise vector to incorporate the errors of marker placement and measurement. The best solution of $\mathbf{R}$ and $\mathbf{T}$ can then be determined by the unit-quaternion approach in the sense of minimizing a least-squares error (ref.~\cite{Eggert1997}). The set of $(\mathbf{R},\mathbf{T})$ determined from motion capture are used as samples for training and test. 

\subsubsection{Soft deformable membrane} For the soft deformable membrane, a parameterization based on B\'ezier is conducted to represent its shape more compactly than the positions of markers. The positions of $7 \times 7$ ($N=49$) markers are used to provide raw data for presenting the shape of a deformed membrane. A surface fitting process is conducted to generate the control points of a B\'ezier surface patch for describing the shape. In general, a B\'ezier surface maps parameters $(u,v)$ to surface point coordinate $\mathbf{p} \in \Re^3$ as
\begin{equation}
    \mathbf{p}(u,v) = \sum_{i=0}^{m}\sum_{j=0}^{n}B_{i,m}(u)B_{j,n}(v)\mathbf{c}_{i,j}
\end{equation}
Where $\{\mathbf{c}_{i,j}\}$ are control points of surface, $B_{i,m}(u)$ and $B_{j,n}(v)$ are \textit{Bernstein Basis Polynomials} defined as
\begin{equation}
B_{i,m}(u)=\tbinom{m}{i}u^{i}(1-u)^{m-i}.
\end{equation}
For this hardware setup, the positions of markers can be captured by the motion capture system. For a marker with position $\mathbf{p}_k$, we can determine its parameters ${(u_{k},v_{k})}$ by the marker's planar coordinate when the membrane is flat -- i.e., before pumping air into the chambers. The control points can be determined by minimizing the following energy that measures the square distances between the real coordinates of markers (captured by cameras) and the positions obtained by surface description 
\begin{equation}
    E= \sum_{k=0}^{N-1} (\mathbf{p}_{k}- \sum_{i=0}^{m}\sum_{j=0}^{n}B_{i,m}(u_{k})B_{j,n}(v_{k})\mathbf{c}_{i,j} )^2.
\end{equation}

\begin{figure}[t]
\centering
\includegraphics[width=1\columnwidth]{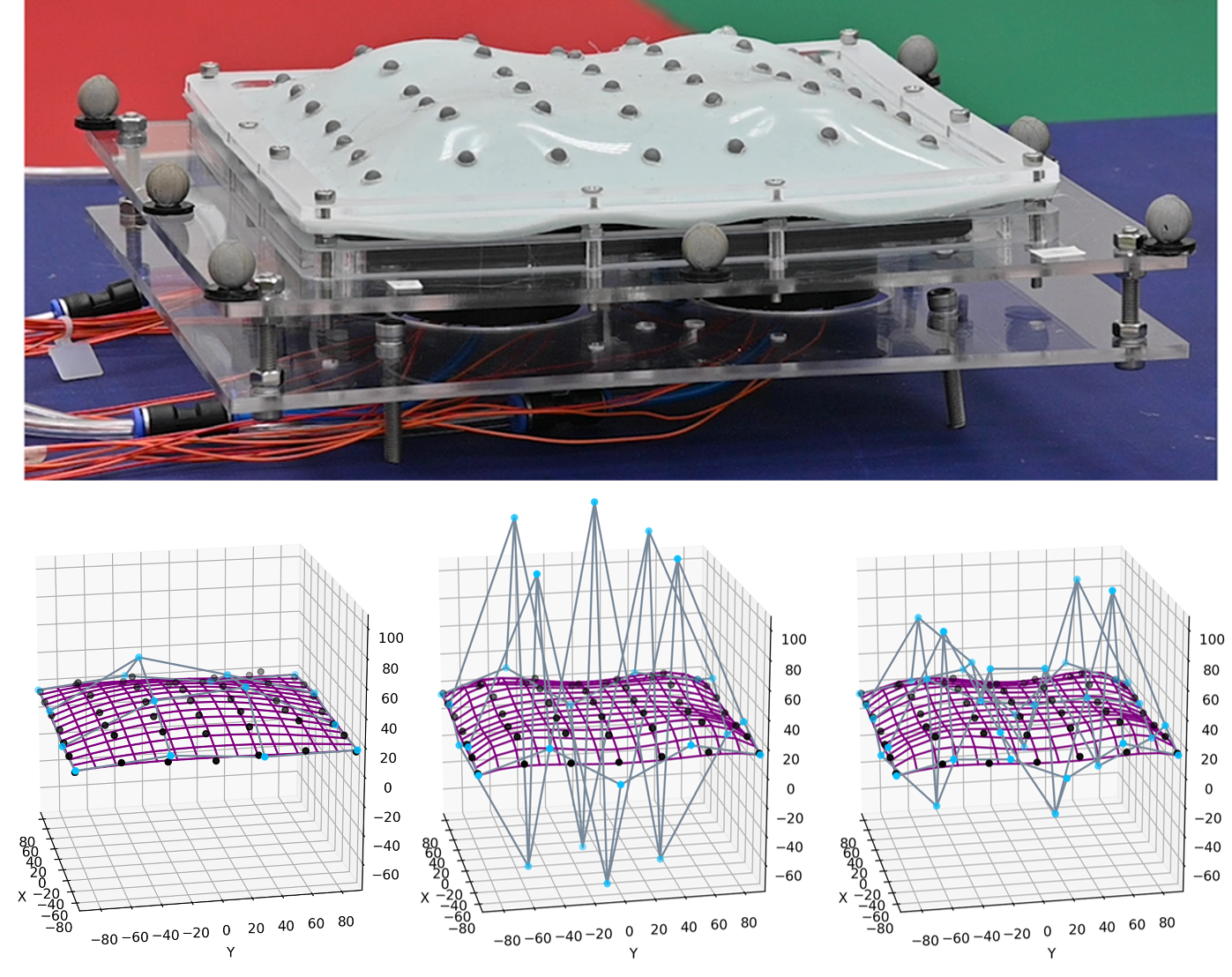}
\caption{Surface fitting for a deformed surface (top row) with 49 markers' positions determined by a motion capture system -- the results by using $4 \times 4$, $5 \times 5$ and $6 \times 6$ control points are given from left to right (bottom row). The black points indicate the measured coordinates of markers, the blue points present the control points obtained by surface fitting and the fitted B\'ezier surfaces are visualized as the purple grids.}\label{fig:fittedsurfaces}
\end{figure}

With the help of B\'ezier surface fitting, the deformable membrane could be expressed as the linear combination of several control points, which removes the redundant information embedded in the positions of markers. Moreover, this compact representation is more robust to noises and outliers. As illustrated in Fig.\ref{fig:fittedsurfaces}, the shape can be represented more accurately when more control points are employed. However, this means that more information needs to be generated from the sensed signals. As a result, a more complex machine learning model is needed which generally also requires to be fed by more training samples. This is challenging for the dataset to be obtained from physical experiments. Therefore, we use smaller number of control points while they can already present the deformed surface accurately enough. Specifically, for this hardware setup of deformable membrane, a B\'ezier surface with $5 \times 5$ control points is employed.

\subsection{Machine Learning}
Machine learning is applied to learn the relationship between the sensor data and the shape oriented parameters. Different learning models are studied in our work to explore the `best' model to be used on different hardware setups. For this purpose, the accuracy is tested on different trained models including a \textit{Long Short Term Memory} (LSTM) network, a \textit{Feedforward Neural Network} (FNN), a \textit{Support Vector Regression} (SVR) and \textit{Multivariate Linear Regression} (MVLR). The network design of these models are presented below.
\begin{itemize}
\item Our network of LSTM has a hidden layer of 50 neurons with $\tanh(\cdot)$ as the activation function. In the output layer, linear function is used for learning the translation and also the control points of B\'ezier surface. Differently $\tanh(\cdot)$ is employed for learning the rotation matrix. This is designed by considering that the rotation has higher nonlinearity than translation and B\'ezier surface, also the output is mapped to the range of $[-1,+1]$. 

\item For the learning model by FNN, we also use a hidden layer of 50 neurons with sigmoid as the activation function. The output layer is designed similar to LSTM -- i.e., a linear function
for the translation and also the control points of B\'ezier surface, and the $\tanh(\cdot)$ function is adopted to learn the rotation matrix for the same reason as LSTM.

\item For SVR, we choose standard \textit{radial basis functions} (RBF) as kernels and use $1.0$ as the $C$-parameter for regularization.

\item MVLR is an ordinary least squares multivariate linear regression. 
\end{itemize}
The preparation of datasets for machine learning is discussed below.

\begin{table}[t]
\centering
\caption{Datasets for Machine Learning}
\begin{tabular}{l|r|r|r} 
\hline
& \multicolumn{3}{c}{Samples \# in Different Datasets} \\
\cline{2-4}
Hardware Setups & Training  & Validation  & Test \\
\hline
\hline
Continuum Joint & $168,251$ & $50,510$ & $23,370$ \\
Deformable Membrane & $21,396$ & $11,503$ & $11,504$ \\
\hline
\end{tabular}
\label{fig:DatasetsML}
\end{table}

For the soft deformable membrane, we obtain the datasets for training, validation and test from the readings from 12 sensors in 10 subsequent time-steps at $1000~\mathrm{Hz}$. That means $12 \times 10 = 120$ readings as input for each prediction. Positions of markers are captured by cameras and converted into control points of B\'ezier surface -- specifically, we generated different control polygons with $4 \times 4$, $5 \times 5$ and $6 \times 6$ to explore the best result. To verify the generality of a learning model's performance, the actuation sequence that was used to generate the dataset of test is different from the actuation sequence used to generate the training dataset. For the soft continuum joint, the datasets are also obtained from the readings from 12 sensors and the captured positions of markers in 10 subsequent time-steps at $1000~\mathrm{Hz}$. Again, $12 \times 10 = 120$ readings as input for each prediction. The positions of markers are converted into a rotation matrix and a translation vector to form a sample. The performance of learning models are evaluated on the dataset captured while external loads are applied. Note that these specific external loads are not applied while generating training data so that the generality of a learning model can be well verified. The total numbers of samples in different datasets and the comparison of mean prediction errors are given in Table~\ref{fig:DatasetsML}.

%% file: secResults.tex
\section{Results}
\label{sec:results}
This section presents the experimental results of applying our approach on the two hardware setups -- the soft continuum joint and the soft deformable membrane. Quantitative analysis has been conducted to verify the performance of our method. 

\subsection{Soft Continuum Joint}
\begin{figure}[t]
   \centering
   \includegraphics[width=.9\columnwidth]{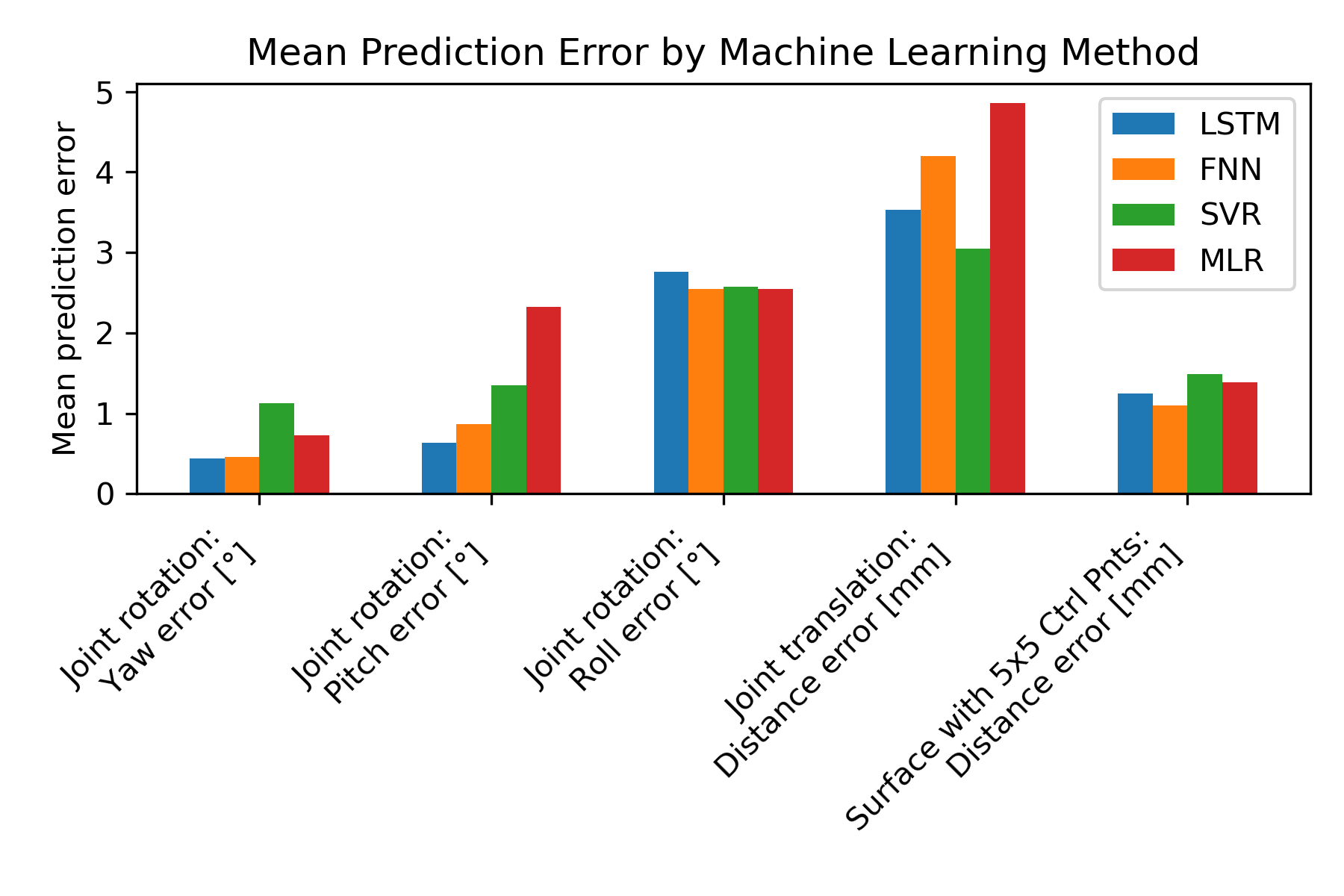}
   \vspace{-15pt}
    \caption{Comparison of the mean prediction errors by using different learning models for the soft continuum joint and the soft deformable membrane. LSTM provides the best performance for predicting transformations of the continuum joint, whereas FNN provides the best performance for the deformable membrane by using $5 \times 5$ control points. Note that the surface error is evaluated as the distances between real and predicted positions of markers.}
\label{fig:bargraphs_ML_models}
\end{figure}

The performance of different machine learning models is compared and given in Fig.~\ref{fig:bargraphs_ML_models} for both the rotation matrix and the translation vector. 

\begin{figure}[t]
   \centering
   \includegraphics[width=0.492\linewidth]{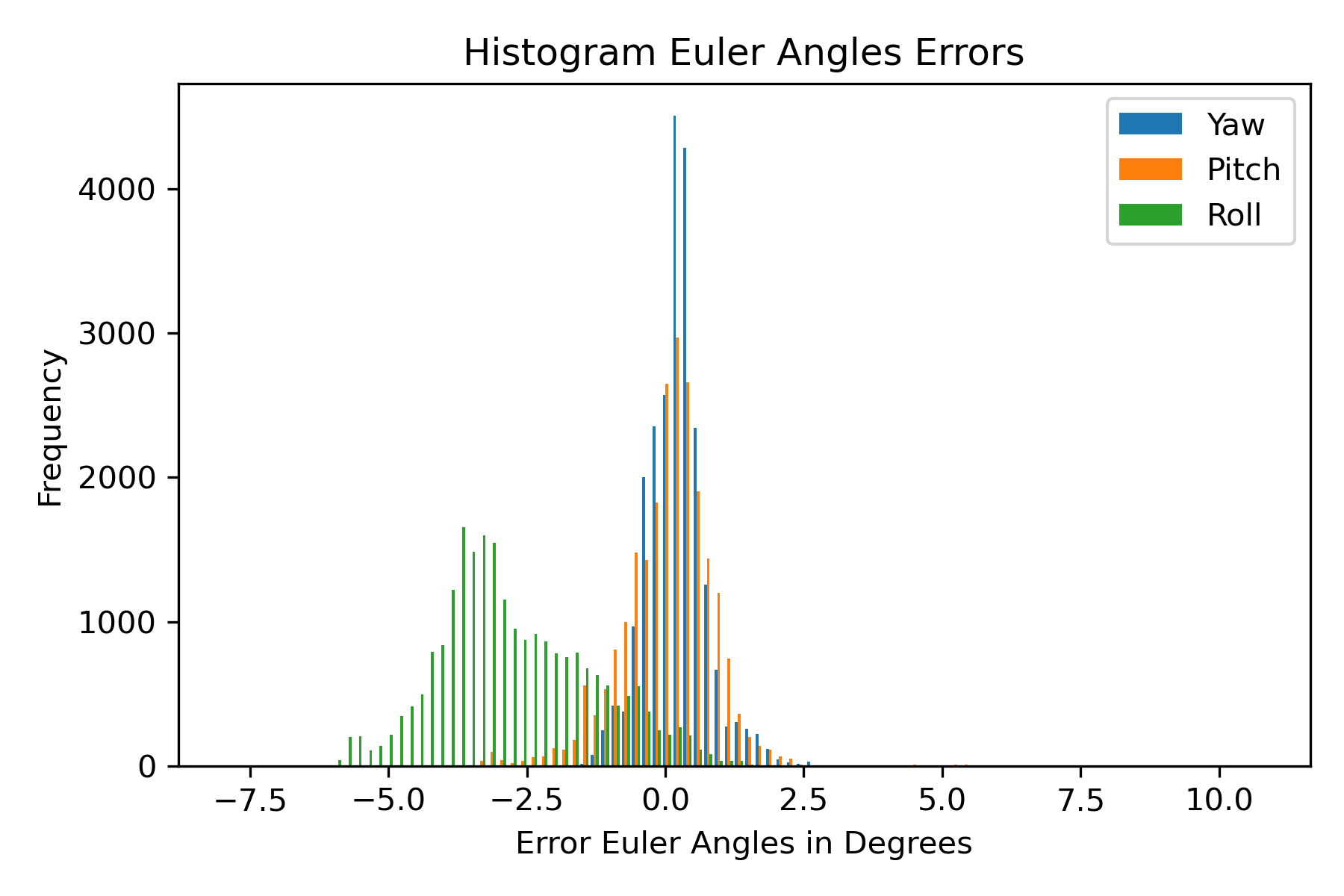}
   \includegraphics[width=0.492\linewidth]{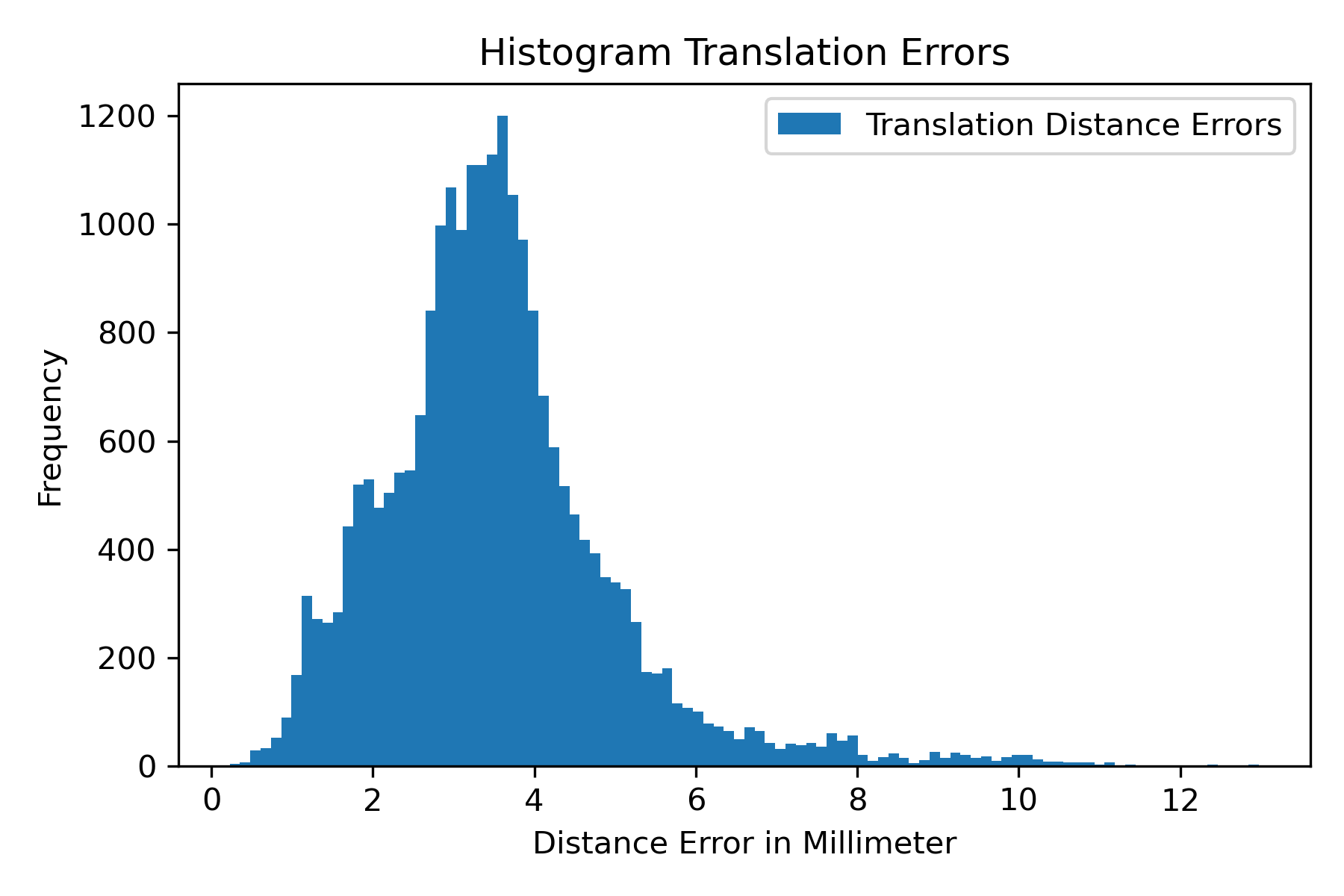}
   \caption{Histograms of the prediction errors for the rotation (left) and the translation (right) of the soft continuum joint, where the prediction is generated by LSTM.}
   \label{fig:continuumJointErrHistogram}
\end{figure}

For providing a more meaningful interpretation for the rotation matrix, the prediction errors of a rotation matrix are translated to Tait-Bryan Euler angles following the $z$-, $y'$- and $x''$-convention (intrinsic rotations) -- referred to as yaw, pitch, and roll respectively. The best performance was achieved by using the LSTM with mean prediction errors as $0.44$, $0.63$, and $2.76$ degrees in yaw, pitch and roll respectively. Prediction error of the translation vector is evaluated by the vector's magnitude. According to the evaluation taken on the test dataset, SVR gives the smallest error as $3.05~\mathrm{mm}$. LSTM's error is $3.53~\mathrm{mm}$, which is comparable to SVR. Therefore, by combining both rotation and translation, the model learned by LSTM provides the best performance. Not only the mean prediction error but also the distribution is shown in Fig.~\ref{fig:continuumJointErrHistogram}.

\begin{figure}[t]
\centering
\includegraphics[width=\columnwidth]{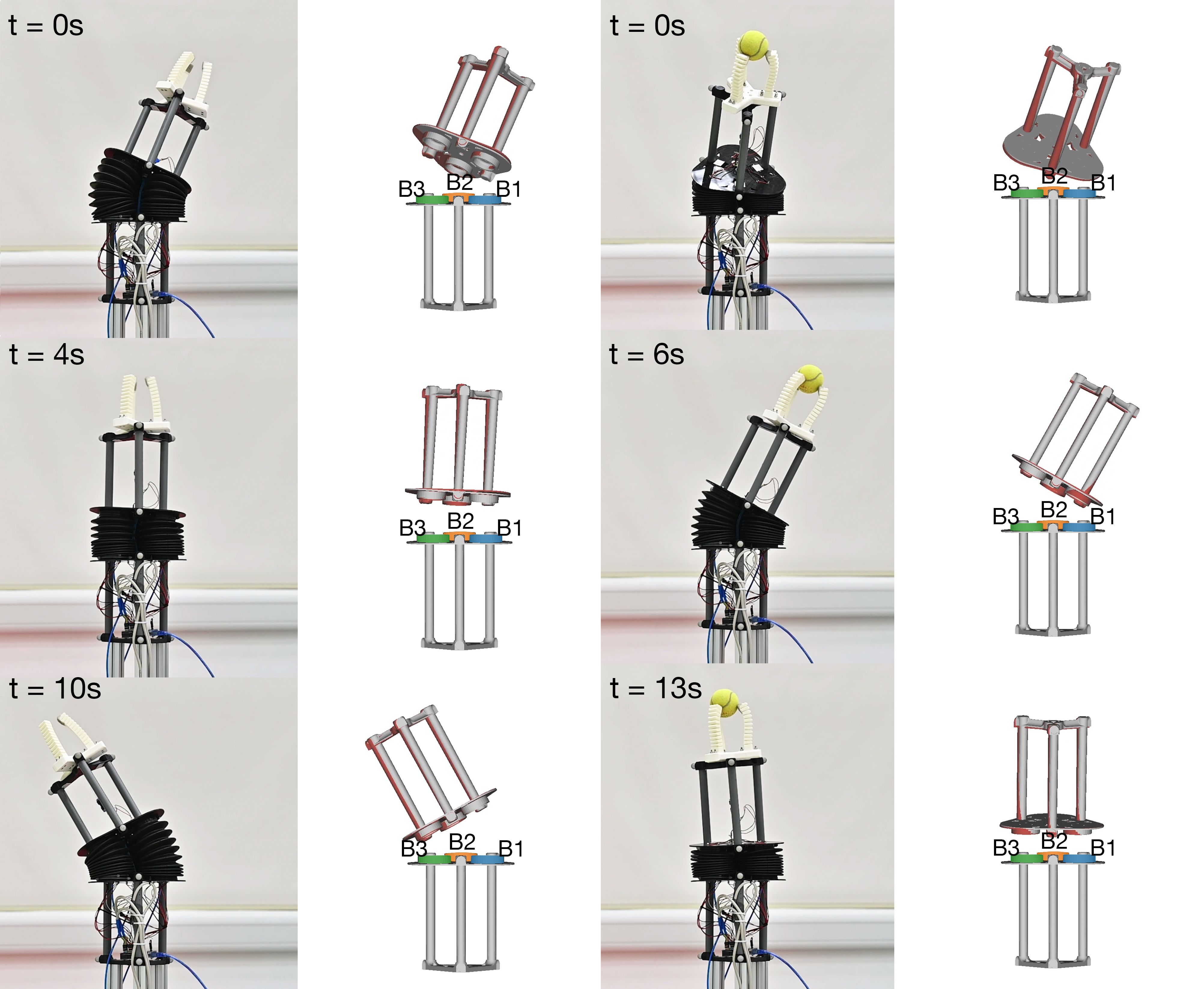}   
\includegraphics[width=0.492\linewidth]{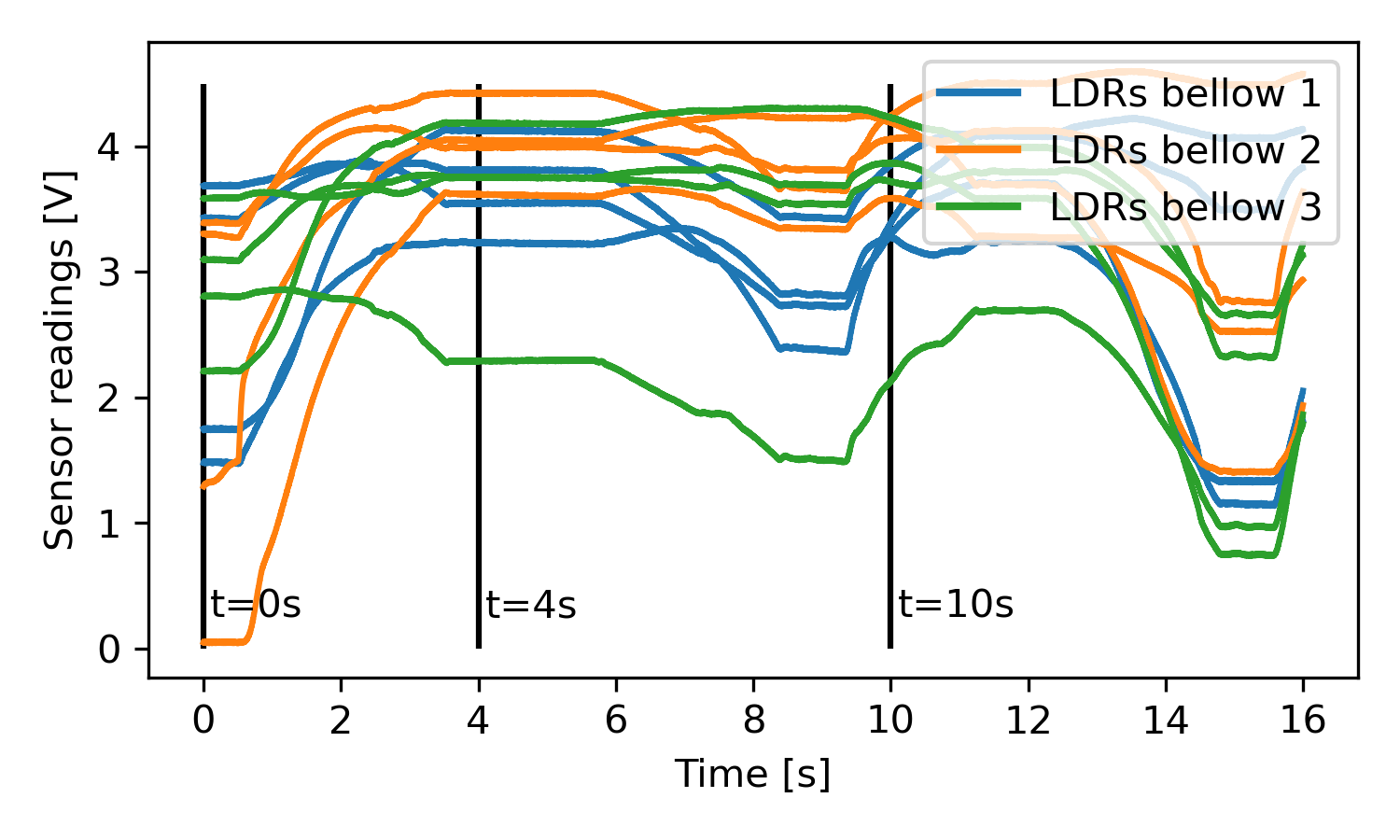}
\includegraphics[width=0.492\linewidth]{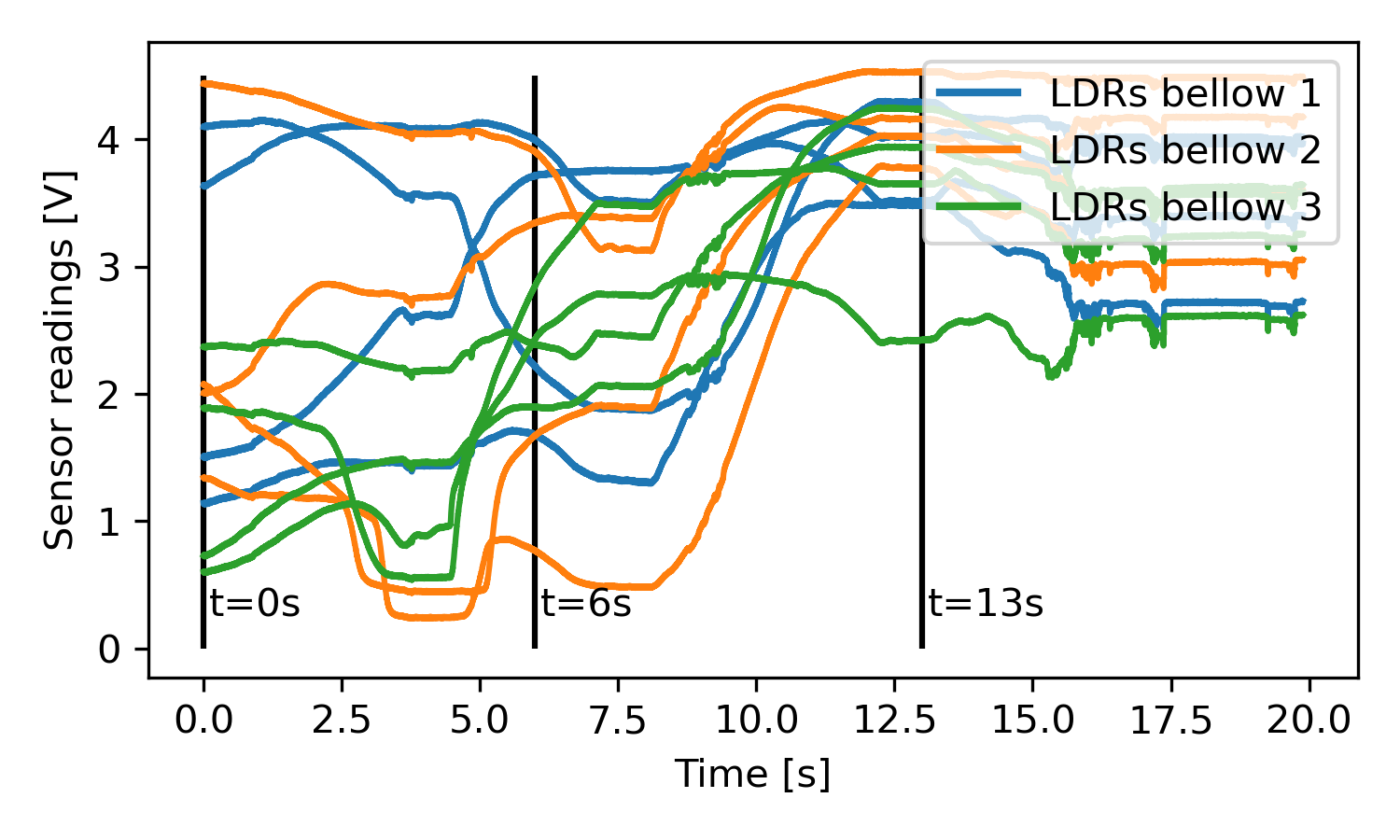}
\caption{Comparison of the rotation and translation predicted by LSTM on the continuum joint and the actual rotation and translation (obtained from motion capture) for a sequence without external load (left) and with external load (right). The transformation obtained from motion capture is displayed in red, whereas the reconstructed transformation from LSTM prediction is rendered in gray. Sensor readings of the LDRs in the different modules throughout the time sequences are shown in the bottom graphs. Bellows and their corresponding set of LDR sensor readings are visualized in matching colors.}
\label{fig:comparisonjoint}
\end{figure}

A side-by-side comparison of the reconstructed joint and the real joint position for a time sequence of 16 seconds is shown in Fig.~\ref{fig:comparisonjoint} and the online video (\url{https://youtu.be/T9wqiIr3s-c}). This reconstruction is based on the predictions obtained from the LSTM model, which demonstrate the capability of accurate prediction regardless of the external load. 

A prediction for the rotation and translation can be generated within $4~\mathrm{ms}$ on a consumer-level device (i.e., a laptop PC with 2.3GHz CPU + 16GB RAM), whereas the calculation of the forwards kinematics is very fast. In practice, the sensor readings can be obtained at the rate of $1000~\mathrm{Hz}$ and each prediction is made by using readings from 10 time-steps. Therefore, we can make a prediction in single-thread computation within every $14~\mathrm{ms}$, which makes it possible to run the reconstruction in real-time (at the rate between $50$ and $70~\mathrm{Hz}$). The speed of visualizing deformed 3D models as shown in Fig.~\ref{fig:comparisonjoint} depends on the mesh density.

\subsection{Soft Deformable Membrane}
\begin{figure}[t]
\centering
\includegraphics[width=\linewidth]{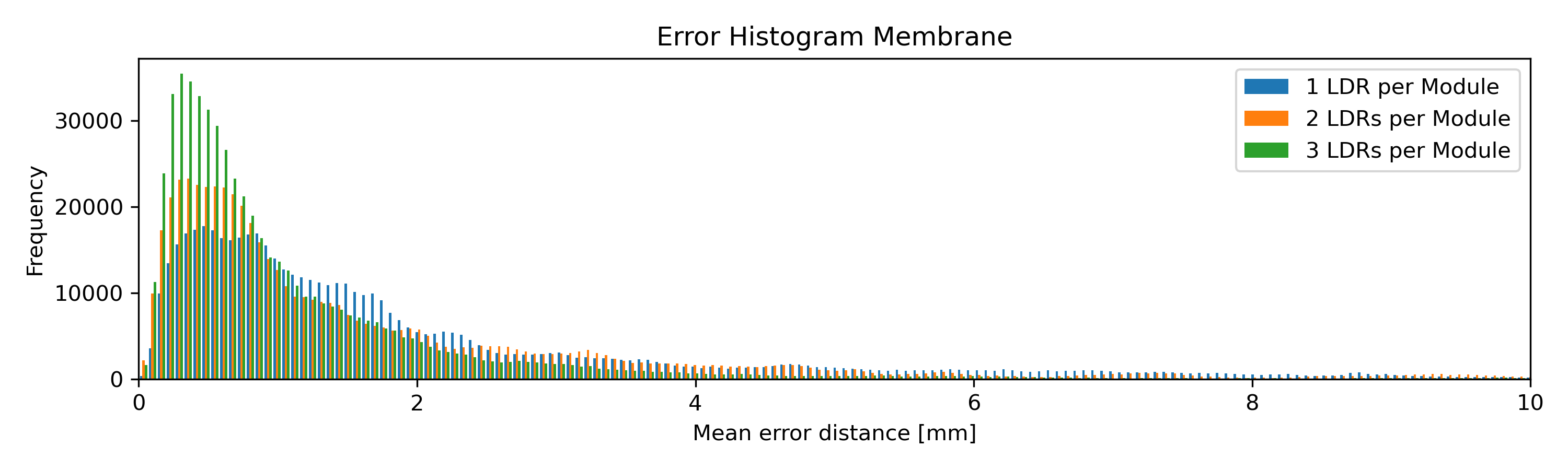}
\footnotesize
\begin{tabular}{r|r|r|r}
\hline
Distance Errors  & 1 LDR & 2 LDRs & 3 LDRs \\
\hline
\hline
Mean Values     &  $2.21 \mathrm{mm}$ & $1.79 \mathrm{mm}$ & $1.15 \mathrm{mm}$ \\ 
Standard Deviation  &  $2.44 \mathrm{mm}$ & $2.06 \mathrm{mm}$ & $1.39 \mathrm{mm}$ \\
\hline
\end{tabular}
\caption{Histogram of distance errors between real and predicted positions of markers on the soft deformable membrane when using different numbers of LDRs in each module. 
}\label{fig:histogramMembrane}
\end{figure}

The prediction error of the soft deformable membrane is indicated in the right of Fig.\ref{fig:bargraphs_ML_models}. For each marker, its $uv$-parameters can used to generate the marker's position on the predicted surface. The distance between the real position of a marker (obtained from motion capture) and its predicted position is employed as a metric to evaluate the error. When comparing the mean, FNN gives the best result with $1.18~\mathrm{mm}$ as the mean of distance errors. An error histogram is given in Fig.~\ref{fig:histogramMembrane}. 
It is also interesting to study the influence of different numbers of sensors. Therefore, we also generate results by the test dataset using only 2 LDRs and 1 LDR per module. Their corresponding error histograms are also shown in Fig.~\ref{fig:histogramMembrane}. It can be observed that the mean distance error by using only one LDR within each module is nearly twice of the error by using 3 LDRs.

\begin{figure}[t]
   \centering
   \includegraphics[width=.6\columnwidth]{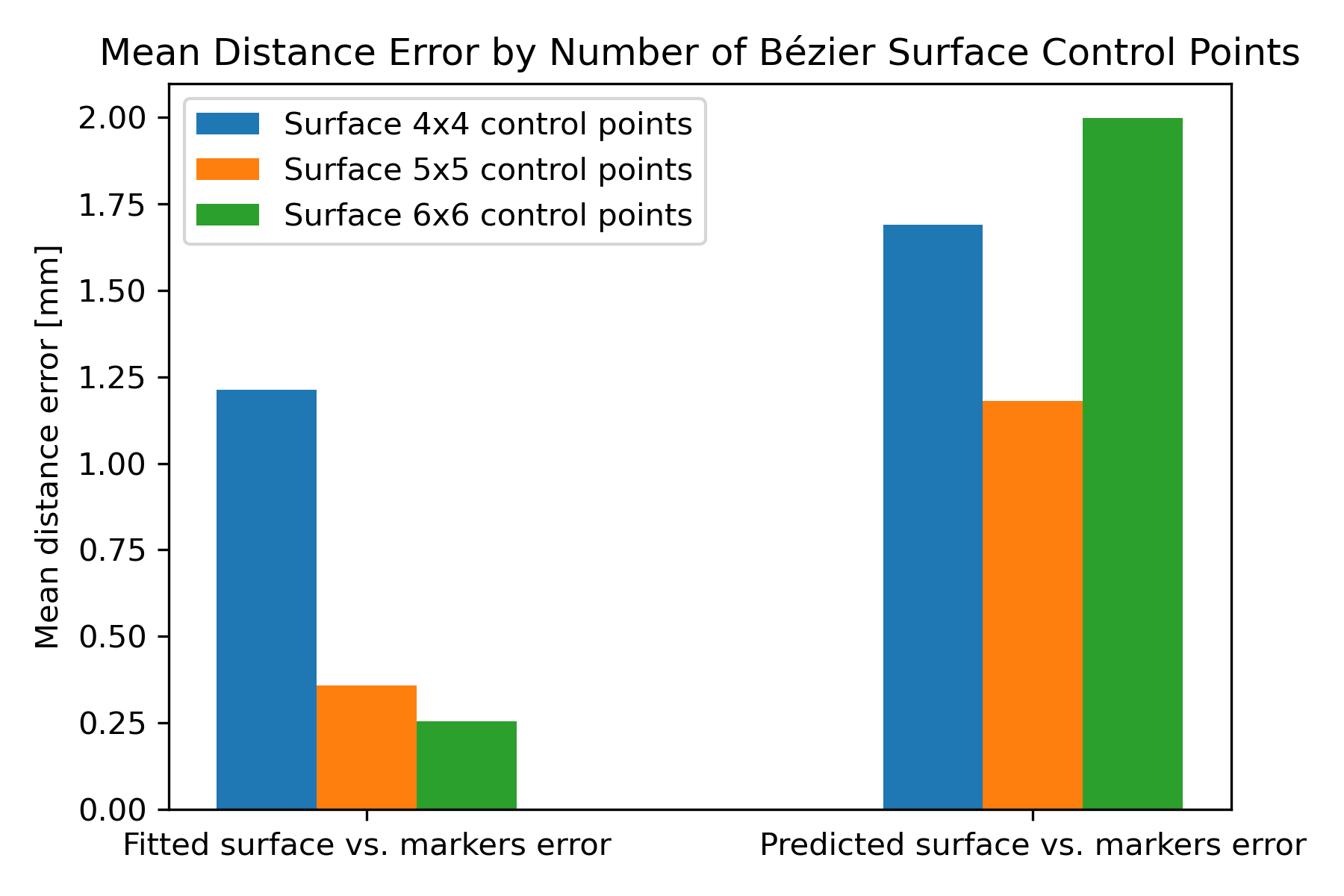}
   \caption{Comparison of the mean distance error for the fitted surface vs. the captured markers, and the predicted surface vs. the captured markers on B\'ezier surfaces with different number of control points.}
   \label{fig:bargraphs_resolution}
\end{figure}

We also study the errors by using different number of control points for B\'ezier fitting. The errors are measured as the distances between the positions of markers and their corresponding points on the resultant surface of fitting. It is obvious that more control points lead to less fitting error (see the left of Fig.~\ref{fig:bargraphs_resolution}). However, the error of surface (with $6 \times 6$ control points) predicted from sensor readings can be larger than the error on a surface with $5 \times 5$ control points (see the right of Fig.~\ref{fig:bargraphs_resolution}). The reason is twofold. First of all, the surface fitting error of $5 \times 5$ is already very close to the error of $6 \times 6$. Secondly, the information from 12 sensors is not very sufficient to predict 36 control points. Therefore, we use $5 \times 5$ control points to represent and reconstruct the soft deformable membrane. 

\begin{figure}
\centering
\includegraphics[width=\columnwidth]{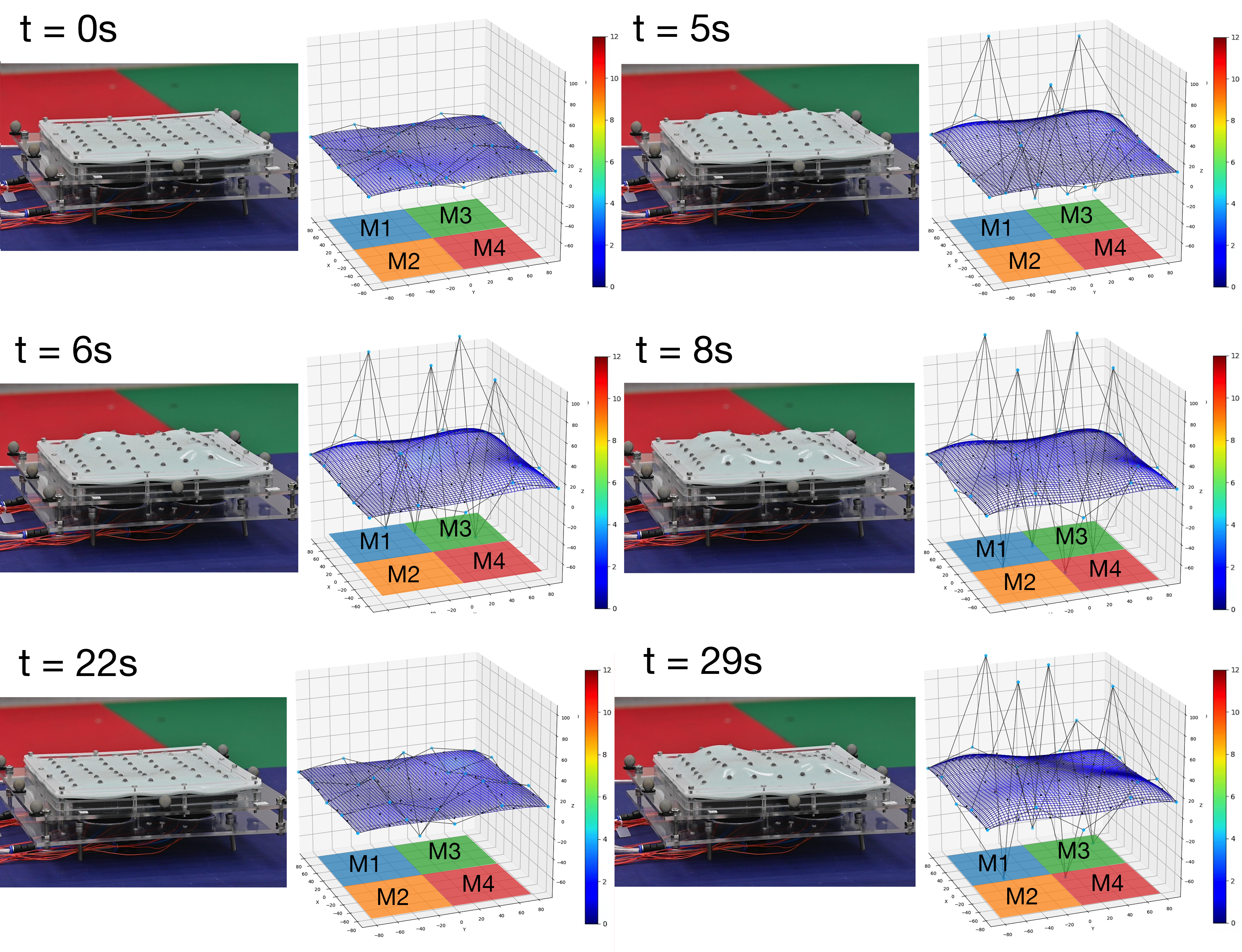}
\includegraphics[width=\columnwidth]{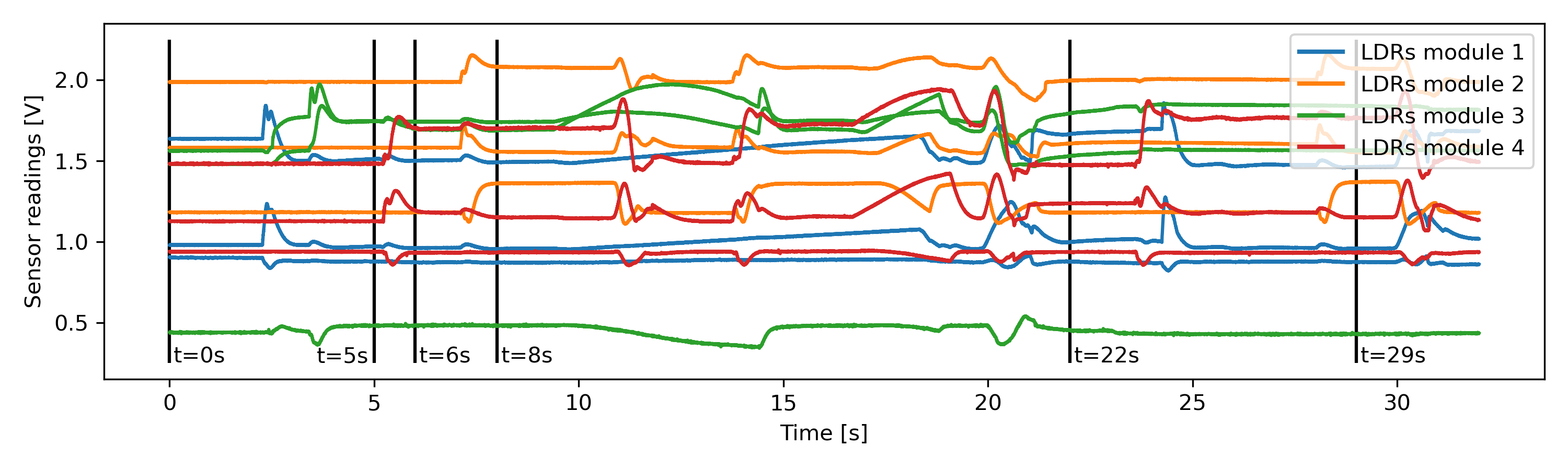}
\caption{Visual comparison of the FNN predicted surface represented by $5 \times 5$ control points and the physically deformed soft membrane during a time sequence of $29$ seconds. 
The positions of markers obtained from motion capture are displayed in black dots, 
whereas their corresponding points on the predicted surface are connected by red line segments. The errors between a predicted surface and a captured surface (by fitting camera captured positions of markers) are visualized as color maps. The sensor readings of the LDRs within the different modules throughout the time sequence are shown in the bottom graph. Modules and their corresponding set of LDR sensor readings are indicated in matching colors. 
   }
   \label{fig:comparisonMembrane}
\end{figure}

A visual comparison of the physically deformed membrane and the reconstructed surface during a period of $29$ seconds can be found in Fig.~\ref{fig:comparisonMembrane} and the online video. The surface is predicted by FNN from the light intensities captured by LDRs in each module. The distance errors, between the surface predicted from sensor readings and the surface generated by fitting camera captured positions of markers, are visualized as color maps. 

A prediction of the control points from sensor readings can be generated within $1~\mathrm{ms}$ on a consumer-level device. Again we make a predication by using the readings from 10 time-steps, which is captured at the rate of $1000~\mathrm{Hz}$. The speed of visualization as shown in Fig.~\ref{fig:comparisonMembrane} strongly depends on the density of the grid -- e.g., a visualization with a $30 \times 30$ grid can be generated within $6~\mathrm{ms}$ using a C++ implementation for visualization. Incorporating all these computations, our system can be operated in real-time at the rate of more than $50~\mathrm{Hz}$.

%% file: Discussion.tex
\section{Conclusion}
\label{sec:discussion}
In this paper, we presented a method to sense and reconstruct 3D deformation on pneumatic soft robots composed of multiple actuators. Our method is based on integrating multiple low-cost sensors inside the chambers of pneumatic actuators and then using machine learning to fuse the captured signals into shape parameters of the soft robots. These shape parameters can be used to efficiently reconstruct the 3D shape of the soft robot.  The sensing and shape prediction pipeline can run at 50 Hz in real-time on a consumer-level device. This is an important step towards the development of more advanced closed-loop control for soft robots. 

Although the method can be applied to a wide range of pneumatic soft robots without modifying their design, the accuracy of proprioception can be further increased by embedding more complex signal generators such as color patterns~\cite{Scharff2018} into the air chambers. Designing the sensors and actuators of the soft robot in concert is therefore expected to yield the best results. The approach discussed in this paper supports such an integrated design approach. Future research could investigate the use of transfer learning to reduce the number of new training samples needed when minor modification is applied to a design.